\documentclass[10pt,twocolumn,letterpaper]{article}

\usepackage[pagenumbers]{cvpr}

\usepackage{colortbl}

\definecolor{cvprblue}{rgb}{0.21,0.49,0.74}
\usepackage[pagebackref,breaklinks,colorlinks,allcolors=cvprblue]{hyperref}

\def\confName{CVPR}
\def\confYear{2027}

\title{Multi-view Random-set Learning}

\author{
{\bfseries
Mengchi Zuo, Tianyang Wang, Cai Xu\thanks{Corresponding author.},
Ziyu Guan, Wei Zhao
}\\[3pt]
{\normalsize School of Computer Science and Technology, Xidian University, Xi'an, China}\\
{\small E-mail: \{mczuo@stu., tianyangwang@stu., cxu@, zyguan@, ywzhao@mail.\}xidian.edu.cn}
}

\begin{document}
\maketitle
\begin{abstract}
Trusted multi-view learning aims to integrate heterogeneous views for accurate and reliable predictions. However, multi-view data often exhibit the view degradation phenomenon: individual views contain only partial discriminative information and cannot independently determine the instance-level label. Requiring such views to produce unambiguous class evidence can distort evidence learning and mislead subsequent fusion. Existing methods addressing this problem have at least one of the following drawbacks: (1) assigning evidence to individual classes without explicitly representing unresolved class groups; (2) projecting composite beliefs onto individual classes before fusion, preventing direct refinement of candidate groups through cross-view set intersections.

To address these limitations, we propose a new Multi-view Random-Set Learning (MRSL) paradigm, which preserves view-specific support for candidate class sets and refines candidate classes through cross-view fusion. Specifically, we quantify class overlap to select a compact collection of sets for each view, avoiding the exponential cost of modeling all class subsets. We then perform set-level fusion by assigning joint support to set intersections, allowing complementary views to narrow the candidate classes before probability conversion. Experiments on synthetic and eight real-world datasets validate the proposed overlap measure and demonstrate that MRSL achieves the highest accuracy on all eight datasets. Compared with the strongest baselines, MRSL improves average accuracy by $4.71\%$ and reduces average expected calibration error by $6.33\%$. All datasets and code will be made publicly available upon acceptance.
\end{abstract}    
\section{Introduction}
\label{sec:intro}


Multi-view learning has found widespread applications, yet many challenges remain in effectively and reliably integrating heterogeneous views. In recent years, trusted multi-view learning has attracted increasing attention by jointly modeling class support and predictive uncertainty for more reliable decisions~\cite{liang2025tmcek,liu2026saef}. Evidential multi-view learning is a representative paradigm~\cite{han2021trusted}, which usually parameterizes Dirichlet distributions with view-specific evidence and fuses the resulting belief masses and uncertainty using a reduced Dempster's combination rule. These approaches have demonstrated promising results across diverse scientific and engineering applications, including automated crystal-system identification~\cite{chen2023automated}, multi-sensor fault diagnosis~\cite{yu2026mscaf}, risk-aware robotic navigation~\cite{cai2024evora}, medical image segmentation~\cite{philps2025uncertainty}, and text-guided protein design~\cite{ge2026protocycle}.

\begin{figure}[t]
    \centering
    \includegraphics[width=\linewidth]
        {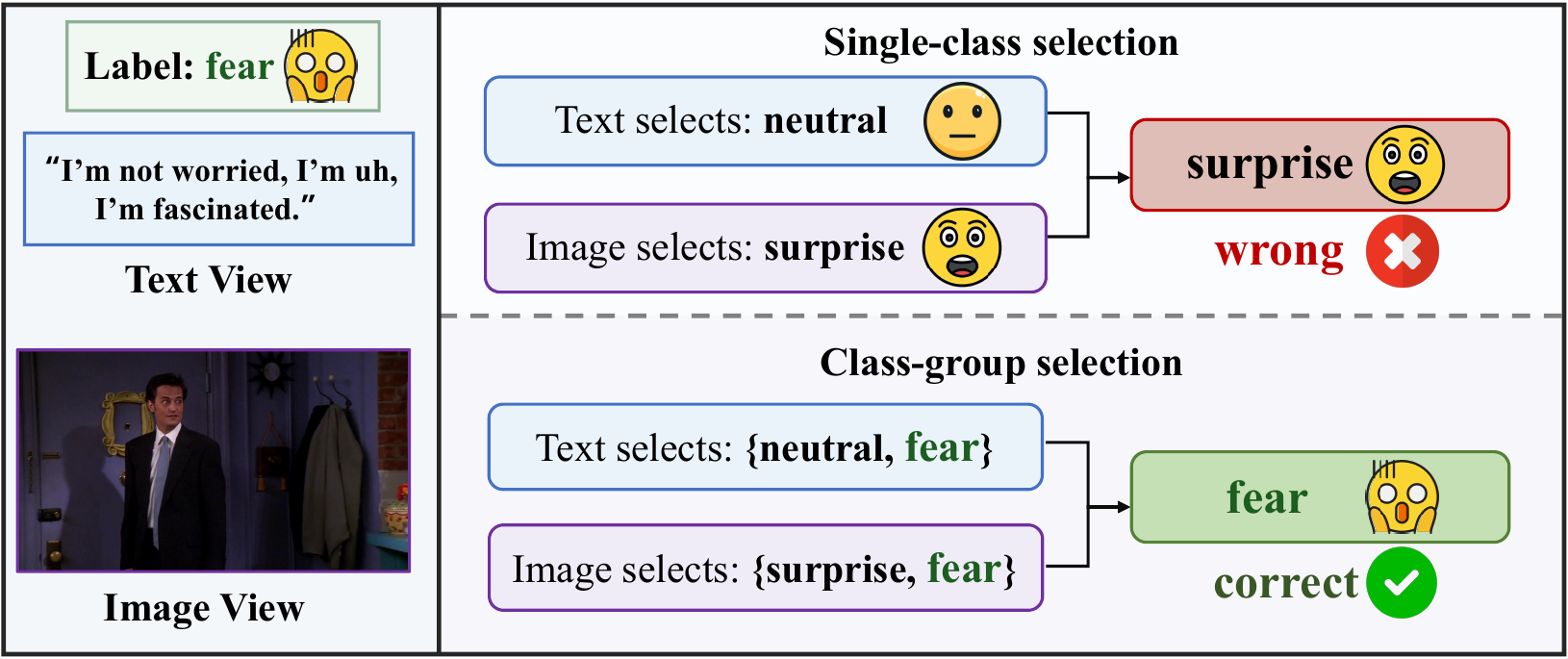}
    \caption{
An illustration of the view degradation phenomenon. Each view provides only partial discriminative information. Forcing a definite class prediction can lead to an incorrect result (top), whereas retaining candidate class groups allows complementary views to preserve \emph{fear} as their shared candidate (bottom).
}
    \label{fig:motivation}
\end{figure}


However, we identify an inherent limitation of existing evidential multi-view learning methods: their difficulty in handling \textit{the view degradation phenomenon}. This phenomenon refers to cases where an individual view contains only partial discriminative information and cannot independently determine the instance-level label. As illustrated in the left and upper parts of Fig.~\ref{fig:motivation}, the text is compatible with both \emph{neutral} and \emph{fear}, while the image is compatible with both \emph{surprise} and \emph{fear}. Although the true label is \emph{fear}, requiring each view to resolve its ambiguity independently can favor an incorrect class and produce misleading support for subsequent fusion. \textit{This creates a mismatch in the evidential multi-view learning paradigm, which encourages each view to produce clear class-specific evidence for the annotated label.} When a view lacks sufficient discriminative information, such supervision can suppress support for other plausible classes and distort view-specific evidence learning. These distortions can propagate through evidence aggregation, compromising the accuracy and reliability of the final decision.

Recently, some trusted multi-view learning methods considering a similar problem have been proposed. Liu et al.~\cite{liu2024dynamic}  dynamically decouple consistent and complementary evidence and relaxe supervision on the complementary component to accommodate semantic vagueness. However, its evidence is still assigned to individual classes, leaving support for unresolved class groups implicit. Hu et al.~\cite{hu2026hyperopinion} introduce hyper-opinions over singleton and composite classes to quantify vagueness and use this information to modulate gradients during training. However, it projects composite beliefs onto individual classes before cross-modal fusion. Consequently, class groups are not explicitly retained during fusion, preventing their direct refinement through cross-view set intersections.


To this end, we propose a new Multi-view Random-Set Learning (MRSL) paradigm. The overall idea is to retain each view's support for candidate class sets and exploit cross-view complementarity to progressively refine the candidate classes for the final prediction. This paradigm introduces two challenges. First, how should candidate sets be selected? For $C$ classes, there are $2^C-1$ nonempty subsets. Modeling all subsets is computationally expensive and may disperse the model's support, hindering learning. To address this challenge, we measure view-specific class overlap using training features and labels. We independently select a small collection of candidate sets associated with each view's ambiguity structure, constructing a compact candidate space. Second, how should fusion be performed? Different views may support overlapping class groups. Aggregating support only for matching sets overlooks relationships between groups, while early probability conversion loses set structure. We therefore assign the joint support of different sets to their intersections, exploiting cross-view complementarity to refine candidate classes while preserving set structure. Experiments on synthetic and real-world data validate our overlap measure and show gains in classification accuracy and probability calibration.

Our main contributions are summarized as follows:
\begin{itemize}
    \item We identify the adverse effects of view
    degradation on trusted multi-view learning and
    propose MRSL, a new set-based framework that preserves candidate-class support when individual views provide insufficient discriminative information.

    \item By quantifying view-specific class overlap
    to select candidate sets and refining their support
    through set-level fusion, MRSL exploits
    complementary information while retaining unresolved ambiguity for confidence estimation, improving both classification performance and confidence calibration.

    \item Experiments on synthetic and real-world data show that MRSL achieves the highest accuracy on all eight real-world datasets, with $4.71\%$ higher average ACC and $6.33\%$ lower average ECE than the strongest baselines. 
\end{itemize}
\section{Related work}
\label{sec:related_work}

\subsection{Trusted multi-view learning}

Trusted multi-view learning aims to achieve accurate predictions together with reliable uncertainty estimates. Evidential learning has become a major paradigm for this goal. It typically transforms view-specific network outputs into class evidence, parameterizes Dirichlet distributions, and derives subjective opinions for uncertainty-aware prediction and fusion~\cite{han2021trusted,xu2024ecml}. Existing studies develop this framework through three main mechanisms. \textbf{Evidence collection} improves the quality and diversity of evidence extracted from multi-view data. DDEF~\cite{shao2024dual} introduces dual-level activation-based evidence extraction, while TEF~\cite{liang2025evolutionary} constructs informative pseudo-views through evolutionary architecture search. \textbf{Uncertainty estimation} refines the relationship between evidence and predictive reliability. RTMC~\cite{zhou2025refining} distinguishes confusion from ignorance, whereas TMCEK~\cite{liang2025tmcek} incorporates evidence distribution characteristics into subjective opinions. \textbf{Opinion aggregation} improves how evidence from different views contributes to the final decision. SAEML~\cite{xu2025beyond} adaptively weights view-specific evidence to preserve informative contributions from weaker views, while GTMC-HOA~\cite{shi2026generalized} employs hierarchical aggregation to exploit both intra-view and inter-view information.

Most existing evidential methods implicitly assume that each view provides sufficient discriminative information and encourage each view to form clear class-specific evidence. However, an individual view may contain only partial information and support a group of plausible classes without distinguishing its members. Requiring such a view to identify a single class may impose discrimination beyond its available information. Consequently, these methods do not adequately address the view degradation phenomenon considered in this work.

\begin{figure*}[t]
    \centering
    \includegraphics[width=0.8\textwidth]
    {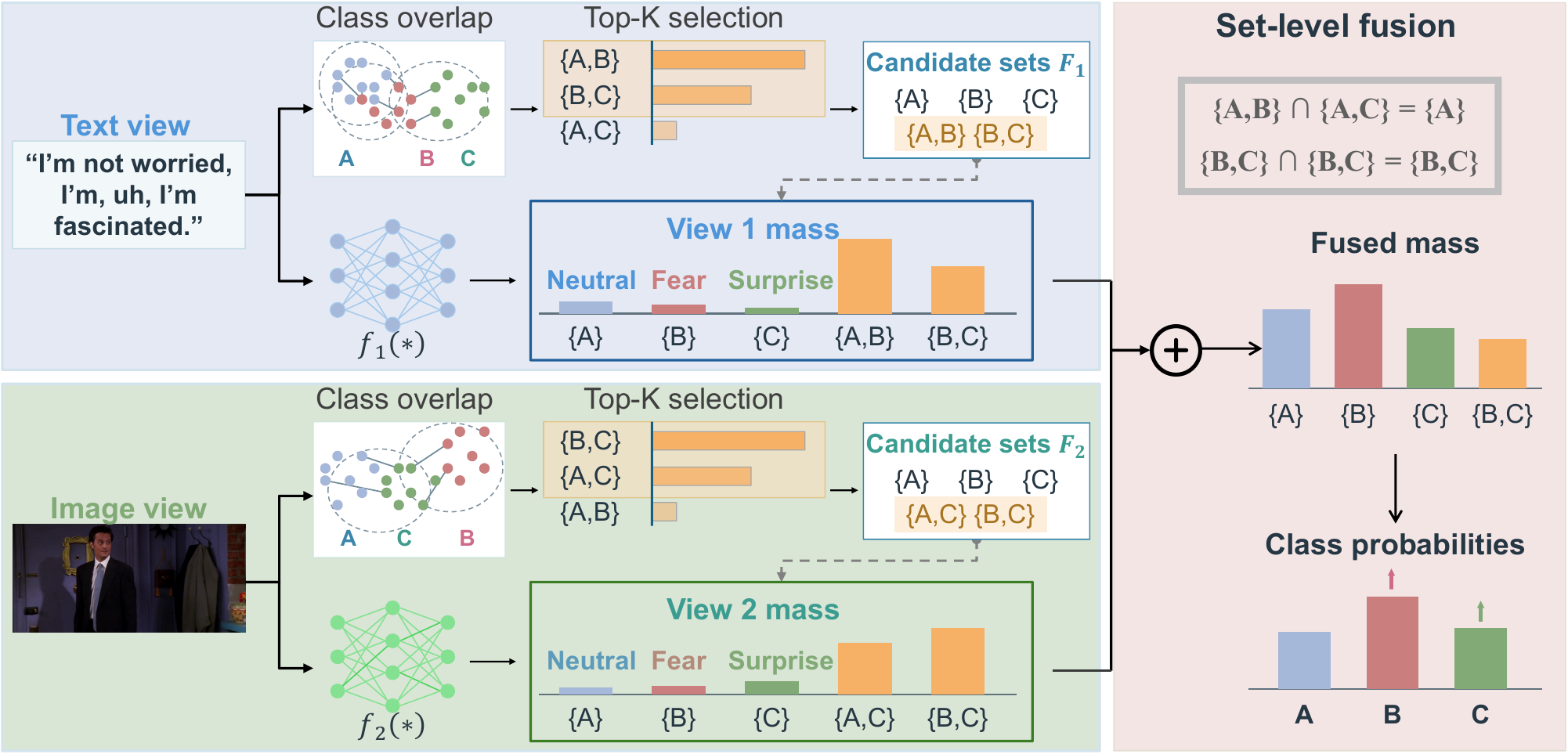}
    \caption{
Overall framework of MRSL. Each view independently selects binary candidate sets based on training-set class-pair overlap and predicts masses over individual classes and the selected candidate sets. DS fusion uses set intersections to refine the candidate classes, and the fused masses are then converted into class probabilities.
}
    \label{fig:rsmc_framework}
\end{figure*}

\subsection{Random-set learning}
\label{sec:random_set_classification}

Random-set learning is an emerging neural learning architecture that represents predictions through belief functions and associated mass assignments over class subsets~\cite{manchingal2025random}. This representation subsumes ordinary probabilistic predictions while retaining support for class groups, providing a flexible framework for representing and combining imprecise information~\cite{shafer1976mathematical,smets1994transferable}. However, set-valued outputs increase computational and learning costs and introduce additional requirements for candidate-subset construction and budget configuration. Recent studies extend random-set outputs to large language models and graph neural networks for token-level and node-level prediction~\cite{mubashar2025randomset,woodley2026randomset}. Related composite-evidence and credal-set representations have also been applied to out-of-distribution detection and ensemble learning~\cite{qu2024hedl,wang2024credal}.


However, these methods are not specifically designed for multi-view learning. Extending random-set learning to this setting introduces two challenges: (1) selecting relevant candidate class subsets for each view within a limited budget, given their different class-overlap structures; (2) jointly learning and fusing support for different subsets across views, so that complementary information can refine candidate classes without losing their set structure.
\section{Method}
\label{sec:method}

\subsection{Problem definition}

Consider a multi-view training set with $N$ instances,
$V$ views, and $C$ classes:
\begin{equation}
    \mathcal{D}
    =
    \left\{
        \left(\{\mathbf{x}_i^{(v)}\}_{v=1}^{V},y_i\right)
    \right\}_{i=1}^{N},
    \label{eq:multiview_dataset}
\end{equation}
where $\mathbf{x}_i^{(v)}\in\mathbb{R}^{d_v}$ is the feature vector of instance $i$ in view $v$, $d_v$ is its dimensionality, and $y_i\in\Omega=\{1,\ldots,C\}$ is its label. Trusted multi-view classification predicts labels and quantifies predictive uncertainty. We predict the most probable class and define uncertainty as one minus its probability. Classification accuracy (ACC) and expected calibration error (ECE) assess classification performance and confidence quality, respectively.

We focus on view degradation caused by insufficient view-specific discriminative information. A view may separate a class group from the rest but fail to distinguish its members. These groups may differ across views, allowing complementary information to refine them. We aim to preserve class-group support during learning and fusion to improve classification performance and confidence quality.

\subsection{Multi-view random-set learning pipeline}
\label{sec:vagueness_quantification}

To address view degradation, we propose Multi-view Random-Set Learning (MRSL). A random set is a set-valued random variable; class subsets with nonzero mass are focal sets. As illustrated in Fig.~\ref{fig:rsmc_framework}, MRSL learns view-specific random-set representations and exploits cross-view complementarity to refine predictions. To construct a compact output space for random-set prediction, we quantify view-specific class overlap and select a small collection of focal sets. For random-set fusion, we assign the joint support of focal sets to their intersections, narrowing the candidate classes while retaining unresolved ambiguity.

\subsubsection{Class overlap quantification and set selection}

Modeling all $2^C-1$ nonempty class subsets is costly and introduces many focal sets unrelated to view-specific ambiguity. We therefore select two-class focal sets corresponding to highly overlapping class pairs to construct a compact output space for random-set prediction. To quantify overlap directly from training features and labels without training an additional classifier, we address two challenges: (1) class imbalance may obscure existing overlap, causing underestimation; (2) outliers may deviate from their own class distribution and lie near other classes, causing overestimation. The resulting class-pair scores guide view-specific focal-set selection, while sample-level overlap information is aggregated into an overall metric characterizing class ambiguity in individual-view and joint feature spaces.

We first construct cross-class connections from local neighborhoods, whose ranges are determined by within-class proximity and constrained by local geometry to reduce outlier effects. To mitigate class imbalance, we assign equal total capacity to each class and distribute it uniformly among its samples. Capacity-constrained maximum matching then yields class-pair overlap scores while limiting repeated sample contributions. Finally, we retain sample participation across class pairs and aggregate it into sample-level scores, which are averaged within each class and then equally across classes to obtain the overall metric.


\paragraph{Local connection.}
To reduce differences in feature scale and dimensionality, we apply z-score standardization using training-set statistics and divide each view's features by $\sqrt{d_v}$, obtaining $\mathbf{z}_i^{(v)}$. To examine whether combining views reduces overlap, we form the joint representation:
\begin{equation}
\mathbf{z}_i^{(V+1)}=\operatorname{concat}\left(\mathbf{z}_i^{(1)},\ldots,\mathbf{z}_i^{(V)}\right).
\end{equation}

In each analysis space $v\in\{1,\ldots,V+1\}$, a fixed distance threshold may fail to accommodate variations in local sampling density. We therefore determine local scales from within-class neighbors. For a class pair $(a,b)$ with $n_a,n_b\geq2$ samples, we set
\begin{equation}
k_t=\left\lceil\frac{n_t-1}{\sqrt{\min(n_a,n_b)-1}}\right\rceil,\qquad t\in\{a,b\},
\end{equation}
which gives the two classes approximately equal within-class neighborhood proportions. For a sample $i$ from class $t$, let $r_i$ denote its distance to the $k_t$-th nearest same-class neighbor, and let $\mathcal{N}_i$ contain the other same-class samples within this radius.

However, an isolated sample may have same-class neighbors concentrated in a distant region, causing $r_i$ to extend across a sparse gap. To limit this expansion, we constrain the radius using neighborhood dispersion. Let $\bar{\mathbf{z}}_i$ denote the mean of the samples in $\mathcal{N}_i$. We define
\begin{equation}
h_i=2\max_{\ell\in\mathcal{N}_i}
\|\mathbf{z}_{\ell}^{(v)}-\bar{\mathbf{z}}_i\|_2,
\;
\rho_i=\min\{r_i,\min_{\ell\in\mathcal{N}_i\cup\{i\}}h_{\ell}\}.
\end{equation}

These constraints jointly bound the radius using the dispersion of the current neighborhood and its members' neighborhoods. The bound $h_i$ alone may remain large when the initial neighborhood contains scattered samples. We therefore take the minimum of $h_\ell$ over $\mathcal{N}_i\cup\{i\}$, incorporating the neighbors' dispersion bounds to conservatively limit neighborhood expansion. All $h_i$ are computed from the initial neighborhoods; when $|\mathcal{N}_i|<2$, we set $h_i=+\infty$.

We connect cross-class samples when each lies within the other's neighborhood defined by $\rho$. Since different local scales may yield one-way coverage, we also allow one-way connections with additional same-class support: if only $i$'s neighborhood covers $j$, at least one of $i$'s nearest same-class neighbors must lie within $j$'s neighborhood. The reverse case is handled symmetrically, and all tied nearest neighbors may provide support. This condition reduces connections supported by a single sample alone. The resulting connection set is denoted by $\mathcal{E}_{ab}^{(v)}$.

\paragraph{Class-pair overlap and set selection.}
After establishing cross-class connections, we further quantify the overlap they reflect. Directly counting connections may be affected by class sizes and repeated sample contributions. We therefore assign equal total capacity to each class and constrain each sample's total contribution across its incident connections to mitigate class imbalance and repeated counting.

For a class pair $(a,b)$, we assign each class unit total capacity and distribute it uniformly among its samples. Let $\pi_{ij}\geq0$ denote the matching weight on an allowed connection $(i,j)\in\mathcal{E}_{ab}^{(v)}$. The weights satisfy $\sum_{j:(i,j)\in\mathcal{E}_{ab}^{(v)}}\pi_{ij}\leq1/n_a$ for every sample $i$ in class $a$ and $\sum_{i:(i,j)\in\mathcal{E}_{ab}^{(v)}}\pi_{ij}\leq1/n_b$ for every sample $j$ in class $b$. Under these constraints, the class-pair overlap is defined as
\begin{equation}
O_{ab}^{(v)}=\max_{\pi\geq0}\sum_{(i,j)\in\mathcal{E}_{ab}^{(v)}}\pi_{ij}.
\end{equation}

We compute $O_{ab}^{(v)}$ using a maximum-flow algorithm, with details provided in the appendix. The unit class capacities ensure $O_{ab}^{(v)}\in[0,1]$, with larger values indicating greater observed class overlap.

Focal sets corresponding to highly overlapping class pairs are suitable for retaining unresolved group-level support. Since overlapping class pairs may differ across views, we select candidate sets independently for each view. For each view $v\in\{1,\ldots,V\}$, we rank all distinct class pairs $a<b$ with positive $O_{ab}^{(v)}$ in descending order and retain at most $K$ corresponding two-class focal sets, forming $\mathcal{C}_v$. Together with all singleton sets, they define the view-specific output space:
\begin{equation}
\mathcal{F}_v=\bigl\{\{1\},\ldots,\{C\}\bigr\}\cup\mathcal{C}_v.
\end{equation}

All overlap scores and candidate sets are determined solely from the training data. The resulting output spaces are fixed before model training and remain unchanged during optimization.

\paragraph{Overall class overlap.}
Class-pair scores do not reveal whether different overlap relations involve the same samples. We therefore retain sample participation before aggregation. For $y_i=a$, participation relative to class $b$ is the fraction of sample capacity used in matching (analogously for class-$b$ samples):
\begin{equation}
s_{ib}^{(v)}(\pi)=n_a\sum_{j:(i,j)\in\mathcal{E}_{ab}^{(v)}}\pi_{ij}.
\end{equation}

Since maximum matchings may yield different participation values, we retain the original connection and capacity constraints, fix the total matching weight at $O_{ab}^{(v)}$, and select a balanced allocation by minimizing
\begin{equation}
\min_{\pi}\quad
\frac{1}{n_a}\sum_{i:y_i=a}\left[s_{ib}^{(v)}(\pi)\right]^2
+
\frac{1}{n_b}\sum_{j:y_j=b}\left[s_{ja}^{(v)}(\pi)\right]^2.
\end{equation}
Strict convexity in participation ensures unique values, denoted by $s_{ib}^{(v)}$, although the edge weights need not be unique. The maximum-flow/minimum-cut procedure is detailed in the appendix.

To emphasize the strongest overlap while retaining other competing classes, we sort each sample's participation across the other $C-1$ classes in descending order as $s_{i,(r)}^{(v)}$ and define
\begin{equation}
s_i^{(v)}=\sum_{r=1}^{C-1}w_r s_{i,(r)}^{(v)},\qquad
w_r=\frac{C}{(C-1)r(r+1)}.
\end{equation}
The weights sum to one; their derivation is provided in the appendix.

To prevent larger classes from dominating, we average first within each class and then equally across classes:
\begin{equation}
G^{(v)}
=
\frac{1}{C}\sum_{a=1}^{C}
\frac{1}{n_a}\sum_{i:y_i=a}s_i^{(v)}.
\label{eq:overall_class_overlap}
\end{equation}
Larger scores indicate greater observed class ambiguity. Comparing $G^{(1)},\ldots,G^{(V)}$ with $G^{(V+1)}$ characterizes changes in overlap after combining views.

\subsubsection{Set-support learning and fusion}
\label{sec:rsmc_framework}

Different views may support different but overlapping class groups. Aggregating support only for matching sets overlooks relationships between groups, while early conversion to class probabilities loses set structure. We therefore let each view predict mass assignments over class subsets and fuse support through set intersections, exploiting cross-view complementarity to narrow the candidate classes.

\paragraph*{View-specific set support.}
 Each view uses an independent network $f_v$ to predict mass assignments over the predefined output space $\mathcal{F}_v$:
\begin{equation}
    \mathbf{M}_i^{(v)}
    =
    f_v(\mathbf{x}_i^{(v)}).
    \label{eq:view_mass_prediction}
\end{equation}
Here, $M_i^{(v)}(S)$ represents support for the class subset $S\in\mathcal{F}_v$, satisfying $M_i^{(v)}(S)\geq0$ and $\sum_{S\in\mathcal{F}_v}M_i^{(v)}(S)=1$.

\paragraph*{Set-level fusion and classification.}
Set intersections refine the candidate range jointly supported by different views, as in $\{1,2\}\cap\{2,3\}=\{2\}$. Accordingly, we use the DS rule~\cite{shafer1976mathematical} to assign products of set masses to their corresponding intersections. For views $u$ and $v$, the fused mass is computed as
\begin{equation}
    M_i^{uv}(S)
    =
    \frac{
        \displaystyle
        \sum_{S_1\cap S_2=S}
        M_i^{(u)}(S_1)M_i^{(v)}(S_2)
    }{
        1-\kappa_i^{uv}
    }.
    \label{eq:set_level_ds_fusion}
\end{equation}
Here, $S\neq\varnothing$, $S_1\in\mathcal{F}_u$, and $S_2\in\mathcal{F}_v$, while $\kappa_i^{uv}$ is the sum of mass products over all set pairs with empty intersections, requiring $\kappa_i^{uv}<1$, with $M_i^{uv}(\varnothing)=0$. Combining all views yields $\mathbf{M}_i=\mathbf{M}_i^{(1)}\oplus\cdots\oplus\mathbf{M}_i^{(V)}$.

The fused result may still support a class group. To obtain final class probabilities, we use the pignistic transformation~\cite{smets1994transferable}, distributing each set's mass equally among its member classes:
\begin{equation}
    p_{i,c}
    =
    \sum_{\substack{
        \varnothing\neq S\subseteq\Omega\\
        c\in S
    }}
    \frac{M_i(S)}{|S|},
    \qquad c\in\Omega,
    \label{eq:pignistic_projection}
\end{equation}
where $|S|$ denotes the number of classes in $S$. The final prediction is $\hat{y}_i=\arg\max_{c\in\Omega}p_{i,c}$.

\subsection{Training objective}
\label{sec:training_objective}


We train MRSL with a joint cross-entropy loss on the fused and view-specific predictions.

Let $p_{i,c}^{(v)}$ denote the class probability obtained from view $v$ using the same probability transformation as in Sec.~\ref{sec:rsmc_framework}. The overall training objective is
\begin{equation}
    \mathcal{L}
    =
    -\frac{1}{N}\sum_{i=1}^{N}
    \left[
        \log p_{i,y_i}
        +
        \frac{\lambda}{V}\sum_{v=1}^{V}
        \log p_{i,y_i}^{(v)}
    \right],
    \label{eq:training_objective}
\end{equation}
where $\lambda\geq0$ controls the weight of the view-specific auxiliary classification losses.
\section{Experiments}
\label{sec:experiments}

\subsection{Experimental setup}
\label{sec:experimental_setup}

We provide a brief overview of the datasets, compared methods and evaluation metrics below. More detailed descriptions and implementation are provided in the appendix.

\paragraph*{Datasets.}
We evaluate MRSL on eight public multi-view datasets: \textbf{Handwritten}, \textbf{Scene}, \textbf{Caltech-6V}, \textbf{MSRCV}, \textbf{Leaves100}, \textbf{Prokaryotic}, \textbf{CUB}, and \textbf{MELD}.


\paragraph*{Compared methods.}
To comprehensively evaluate MRSL, we compare it with representative multi-view classification methods from two categories: (i) \emph{trusted multi-view learning methods}, including \textbf{TMC}~\cite{han2021trusted}, \textbf{ECML}~\cite{xu2024ecml}, \textbf{CCML}~\cite{liu2024dynamic}, \textbf{TMCEK}~\cite{liang2025tmcek}, \textbf{RCMCL}~\cite{hu2026rcmcl}, \textbf{TUNED}~\cite{huang2025tuned}, \textbf{HOVQ}~\cite{hu2026hyperopinion}, and \textbf{ETF}~\cite{lu2025trust}; and (ii) \emph{feature-level fusion methods}, including \textbf{UVF}~\cite{lv2026uvf}, \textbf{AssoDMVC}~\cite{liang2025assodmvc}, and \textbf{PDF}~\cite{cao2024pdf}.




\paragraph*{Evaluation metrics.}
We report classification accuracy (ACC) and expected calibration error (ECE)~\cite{guo2017calibration}. ECE uses 15 equal-width bins. The eight trusted multi-view learning methods baselines use confidence $1-u$, where $u$ denotes each method's native predictive uncertainty computed from evidence strength. Stronger evidence corresponds to lower uncertainty and hence higher confidence. The remaining methods, including MRSL, use the maximum predicted class probability as confidence.

\subsection{Experiments on synthetic data}
\label{sec:synthetic_study}

We conduct synthetic experiments to address two questions:

\textbf{RQ1:} Can MRSL effectively learn and combine support for ambiguous class groups to improve joint classification?

\textbf{RQ2:} Can the overall class-overlap score $G^{(v)}$ (Eq.~\ref{eq:overall_class_overlap}) capture class overlap and cross-view complementarity?

These questions require known and controllable overlap structures, which are generally unavailable in real-world datasets. We therefore construct synthetic data with adjustable overlap levels, with detailed generation procedures provided in the appendix.

\subsubsection{Strategy effectiveness}
\label{sec:synthetic_classification}

\begin{figure}[t]
    \centering
    \includegraphics[width=\linewidth]{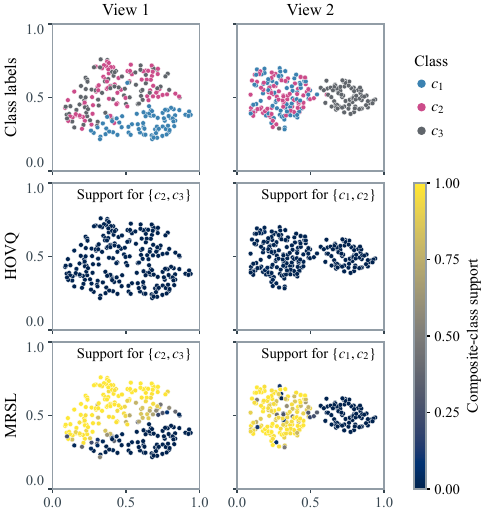}
    \caption{Ground-truth classes (top) and composite-class support of HOVQ (middle) and MRSL (bottom) on the same synthetic test instances. Columns correspond to views~1 and~2.}
    \label{fig:synthetic_data}
    \label{fig:synthetic_composite_support}
\end{figure}

We construct a two-view synthetic dataset with 1,200 instances from three classes and use an 80/20 train/test split. Classes $c_2$ and $c_3$ are difficult to distinguish in view~1, while $c_1$ and $c_2$ are difficult to distinguish in view~2, producing complementary class-overlap structures.

\paragraph*{Classification and calibration performance.}
Table~\ref{tab:synthetic_results} shows that MRSL achieves the highest ACC ($95.17\%$) and the lowest ECE ($3.08\%$), improving ACC over CCML by 4.17 percentage points. Compared with HOVQ, the classification gain is concentrated on $c_2$, whose accuracy increases from $76.25\%$ to $91.50\%$, while the two methods obtain the same mean accuracies on $c_1$ and $c_3$.

\begin{table}[t]
    \centering
    \small
    \setlength{\tabcolsep}{4pt}
    \caption{Classification and overlapping-group detection results on synthetic data (\%, mean $\pm$ standard deviation). A dash indicates unavailable composite-class support.}
    \label{tab:synthetic_results}
    \begin{tabular}{lccc}
        \toprule
        Model & ACC $\uparrow$ & ECE $\downarrow$ & AUROC $\uparrow$ \\
        \midrule
        CCML & $91.00\pm1.63$ & $17.62\pm1.98$ & --- \\
        HOVQ & $90.08\pm1.12$ & $48.14\pm1.09$ & $95.19\pm1.58$ \\
        \textbf{MRSL} & $\mathbf{95.17\pm0.48}$ & $\mathbf{3.08\pm0.47}$ & $\mathbf{98.75\pm0.51}$ \\
        \bottomrule
    \end{tabular}
\end{table}

\begin{table*}[t]
    \centering
    \small
    \setlength{\tabcolsep}{2.5pt}
    \renewcommand{\arraystretch}{1.12}
    \caption{Classification accuracy on eight datasets (\%, mean $\pm$ standard deviation). Best and second-best means are bold and underlined.}
    \label{tab:real_acc_draft}
    \resizebox{\textwidth}{!}{%
    \begin{tabular}{@{}l*{8}{c}@{}}
        \toprule
        Model & Scene & Handwritten & Caltech-6V & MSRCV & Leaves100 & Prokaryotic & CUB & MELD \\
        \midrule
        TMC & $66.44\pm1.15$ & $97.25\pm1.25$ & $93.85\pm0.75$ & $95.24\pm2.92$ & $87.56\pm1.80$ & $56.76\pm0.00$ & $92.83\pm2.40$ & $61.39\pm0.22$ \\
        ECML & $66.13\pm1.37$ & $97.05\pm1.25$ & $93.72\pm0.71$ & $95.71\pm2.61$ & $91.81\pm1.31$ & $69.01\pm3.29$ & $92.67\pm3.08$ & $61.66\pm0.12$ \\
        CCML & $66.24\pm1.27$ & $96.65\pm1.11$ & $94.06\pm0.53$ & $92.38\pm3.53$ & $86.00\pm2.02$ & $69.01\pm5.08$ & $93.33\pm2.12$ & $60.37\pm0.14$ \\
        TMCEK & $70.28\pm1.03$ & $97.20\pm0.99$ & $92.34\pm0.48$ & $79.52\pm5.73$ & $92.75\pm1.52$ & $64.86\pm2.30$ & $92.33\pm2.46$ & $62.08\pm0.11$ \\
        RCMCL & $71.24\pm1.41$ & $97.65\pm1.07$ & $93.51\pm0.66$ & $\underline{97.62\pm1.68}$ & $96.25\pm1.38$ & $69.91\pm3.52$ & $92.50\pm2.83$ & $59.47\pm0.58$ \\
        TUNED & $70.93\pm1.66$ & $97.85\pm0.68$ & $94.06\pm0.50$ & $86.67\pm2.13$ & $89.44\pm1.97$ & $\underline{84.86\pm4.70}$ & $89.83\pm2.53$ & $\underline{62.32\pm0.63}$ \\
        HOVQ & $72.35\pm1.59$ & $97.70\pm0.82$ & $89.79\pm0.45$ & $93.81\pm3.61$ & $73.50\pm2.47$ & $67.75\pm1.34$ & $93.50\pm2.79$ & $61.27\pm0.24$ \\
        ETF & $65.48\pm0.69$ & $96.65\pm0.99$ & $94.35\pm0.53$ & $95.24\pm3.37$ & $96.62\pm0.81$ & $56.76\pm0.00$ & $\underline{93.67\pm1.73}$ & $60.36\pm0.51$ \\
        UVF & $72.37\pm3.78$ & $98.40\pm0.45$ & $\underline{95.44\pm0.27}$ & $90.00\pm4.58$ & $\underline{99.56\pm0.52}$ & $78.92\pm2.75$ & $92.00\pm1.51$ & $58.64\pm4.92$ \\
        AssoDMVC & $\underline{79.67\pm0.87}$ & $\underline{98.50\pm0.88}$ & $94.56\pm1.12$ & $91.43\pm3.19$ & $95.94\pm1.15$ & $81.26\pm3.57$ & $93.17\pm2.31$ & $55.23\pm0.92$ \\
        PDF & $71.88\pm1.40$ & $98.05\pm0.21$ & $94.98\pm0.57$ & $95.24\pm2.92$ & $97.75\pm0.56$ & $58.02\pm1.21$ & $93.17\pm1.81$ & $48.21\pm3.40$ \\
        \midrule
        MRSL & $\mathbf{82.50\pm1.22}$ & $\mathbf{98.85\pm0.58}$ & $\mathbf{95.90\pm0.35}$ & $\mathbf{99.52\pm1.06}$ & $\mathbf{99.75\pm0.14}$ & $\mathbf{91.53\pm2.43}$ & $\mathbf{94.67\pm1.62}$ & $\mathbf{64.70\pm1.69}$ \\
        \bottomrule
    \end{tabular}%
    }
\end{table*}

\paragraph*{Random-set representations and complementary information.}
Figure~\ref{fig:synthetic_composite_support} compares the composite-class support of HOVQ and MRSL in the two views. Compared with HOVQ, MRSL assigns higher support to the overlapping class group $\{c_2,c_3\}$ in view~1 and $\{c_1,c_2\}$ in view~2, while keeping the corresponding support low for relatively clear groups.

To quantify this discrimination, we treat the overlapping class group in each view as positive and the remaining class as the reference group. The detection score is the total native composite-class support. AUROC is computed separately for each view and then averaged across the two views. MRSL achieves an AUROC of $98.75\%$, compared with $95.19\%$ for HOVQ, further supporting its ability to distinguish overlapping from relatively clear groups.


These results indicate that MRSL learns support for view-specific ambiguous class groups and achieves higher accuracy in joint classification with complementary views, particularly for class $c_2$, which is difficult to distinguish in either individual view.

\subsubsection{Class overlap and complementarity}
\label{sec:synthetic_overlap_validation}

We evaluate whether the overall class-overlap score $G^{(v)}$ defined in Eq.~\ref{eq:overall_class_overlap} captures changes in joint-space class overlap and the overlap reduction enabled by cross-view complementarity. For the two-view data in this experiment, the joint-space score is denoted by $G^{(3)}$, corresponding to $G^{(V+1)}$ with $V=2$.

We construct two views using three-class Gaussian synthetic data and control the density overlap coefficient $q\in\{0.01,0.25,0.50,0.75,1\}$ by adjusting class-mean distances. Under the same-pair condition, both views overlap $c_2$ and $c_3$. Under the complementary condition, view~1 overlaps $c_2$ and $c_3$, whereas view~2 overlaps $c_1$ and $c_2$. The two conditions have matched single-view overlap, allowing us to examine how different cross-view relations affect joint overlap.

\begin{table*}[t]
    \centering
    \small
    \setlength{\tabcolsep}{2.5pt}
    \renewcommand{\arraystretch}{1.12}
    \caption{Calibration results on eight datasets (ECE in \%, mean $\pm$ standard deviation). Best and second-best means are bold and underlined.}
    \label{tab:real_ece_draft}
    \resizebox{\textwidth}{!}{%
    \begin{tabular}{@{}l*{8}{c}@{}}
        \toprule
        Model & Scene & Handwritten & Caltech-6V & MSRCV & Leaves100 & Prokaryotic & CUB & MELD \\
        \midrule
        TMC & $46.21\pm1.13$ & $15.89\pm1.32$ & $11.72\pm0.78$ & $20.42\pm2.55$ & $81.51\pm1.85$ & $\underline{8.82\pm1.71}$ & $39.16\pm1.92$ & $17.89\pm1.43$ \\
        ECML & $54.28\pm1.36$ & $60.21\pm1.33$ & $50.98\pm0.92$ & $64.40\pm2.50$ & $88.44\pm1.34$ & $36.34\pm3.47$ & $45.71\pm2.58$ & $33.16\pm1.11$ \\
        CCML & $46.86\pm1.26$ & $47.65\pm1.24$ & $45.92\pm0.79$ & $41.92\pm2.95$ & $81.48\pm2.03$ & $19.67\pm2.89$ & $36.88\pm2.39$ & $17.13\pm1.47$ \\
        TMCEK & $54.38\pm1.03$ & $64.74\pm1.05$ & $48.88\pm0.95$ & $27.29\pm6.05$ & $87.66\pm1.57$ & $25.70\pm6.74$ & $38.68\pm2.22$ & $18.69\pm0.70$ \\
        RCMCL & $39.76\pm1.17$ & $26.52\pm1.08$ & $26.49\pm0.54$ & $23.47\pm1.52$ & $83.98\pm1.33$ & $10.59\pm5.08$ & $38.49\pm2.23$ & $\mathbf{6.14\pm1.21}$ \\
        TUNED & $54.58\pm1.52$ & $53.49\pm1.30$ & $58.09\pm3.67$ & $40.56\pm1.24$ & $76.94\pm2.14$ & $40.27\pm6.08$ & $58.45\pm3.05$ & $36.98\pm3.45$ \\
        HOVQ & $58.54\pm1.49$ & $56.55\pm0.98$ & $69.30\pm1.01$ & $51.01\pm3.52$ & $69.13\pm2.50$ & $29.69\pm2.55$ & $73.61\pm2.56$ & $30.24\pm0.87$ \\
        ETF & $47.88\pm0.66$ & $27.33\pm1.06$ & $18.34\pm0.80$ & $32.44\pm2.72$ & $91.21\pm0.83$ & $20.36\pm2.11$ & $46.08\pm1.90$ & $19.32\pm3.73$ \\
        UVF & $19.31\pm2.47$ & $\underline{1.13\pm0.49}$ & $\mathbf{2.16\pm0.94}$ & $\underline{10.50\pm4.69}$ & $\underline{1.27\pm0.25}$ & $13.33\pm3.04$ & $\underline{5.19\pm1.52}$ & $34.51\pm4.07$ \\
        AssoDMVC & $\mathbf{7.72\pm0.69}$ & $8.72\pm2.65$ & $6.24\pm1.26$ & $22.62\pm2.62$ & $3.40\pm0.23$ & $13.52\pm0.46$ & $12.45\pm1.15$ & $35.63\pm0.79$ \\
        PDF & $15.93\pm1.74$ & $3.97\pm0.60$ & $3.56\pm0.75$ & $19.54\pm1.20$ & $39.27\pm0.69$ & $28.07\pm2.93$ & $7.13\pm1.52$ & $35.37\pm6.08$ \\
        \midrule
        MRSL & $\underline{12.09\pm1.16}$ & $\mathbf{1.04\pm0.46}$ & $\underline{3.54\pm0.38}$ & $\mathbf{0.36\pm0.70}$ & $\mathbf{0.27\pm0.15}$ & $\mathbf{7.43\pm1.89}$ & $\mathbf{4.37\pm1.31}$ & $\underline{7.68\pm2.40}$ \\
        \bottomrule
    \end{tabular}%
    }
\end{table*}

We use the normalized Bayes error of the known joint distributions as a theoretical reference:
\begin{equation}
    \widetilde{R}^*(q)
    =
    \frac{
        1-\frac{1}{3}\displaystyle\int
        \max_{c\in\{1,2,3\}}
        f_c(\mathbf{x};q)\,\mathrm{d}\mathbf{x}
    }{2/3},
    \label{eq:normalized_bayes_error}
\end{equation}
where $f_c(\mathbf{x};q)$ is the class-conditional density of class $c$ in the joint feature space. The denominator $2/3$ is the Bayes error when all three equally likely classes fully overlap. Numerical evaluation details are provided in the appendix.

\begin{figure}[t]
    \centering
    \includegraphics[width=\linewidth]{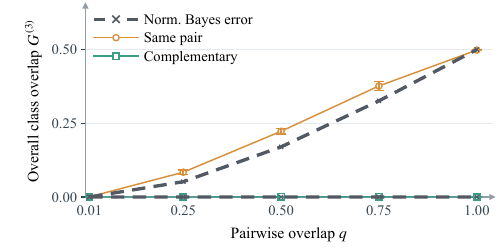}
    \caption{Joint-space overall class-overlap score $G^{(3)}$ under same-pair and complementary conditions. Dashed lines show normalized Bayes error; error bars indicate mean $\pm$ standard deviation.}
    \label{fig:synthetic_G_validation}
\end{figure}

Figure~\ref{fig:synthetic_G_validation} compares the joint-space overall class-overlap score $G^{(3)}$ with the corresponding theoretical references.

\paragraph*{Class overlap.}
Under the same-pair condition, the same class pair is difficult to distinguish in both views. As $q$ increases, $G^{(3)}$ rises and reaches $0.498\pm0.001$ at $q=1$. Its trend agrees with the theoretical reference, indicating that the score captures changes in joint-space class overlap.

\paragraph*{Cross-view complementarity.}
At matched single-view overlap, the complementary condition yields near-zero joint-space scores, whereas the same-pair condition exhibits increasing joint overlap. In the complementary condition, each view provides discriminative information missing from the other, allowing the jointly observed features to distinguish classes that overlap in an individual view. The results agree with the corresponding theoretical references, indicating that the score also reflects the reduction in joint overlap enabled by cross-view complementarity.

\subsection{Experiments on real-world data}
\label{sec:real_data_results}

\subsubsection{Performance comparison}
\label{sec:real_performance_comparison}


\paragraph*{Classification performance.}
Table~\ref{tab:real_acc_draft} compares classification accuracy on eight datasets. MRSL achieves the highest mean ACC on all eight datasets, with an eight-dataset average of $90.93\%$, compared with $86.22\%$ for AssoDMVC, the strongest baseline by average ACC, yielding an improvement of 4.71 percentage points.

\paragraph*{Calibration performance.}
Table~\ref{tab:real_ece_draft} compares calibration results. MRSL achieves the lowest mean ECE on five datasets and ranks second on Scene, Caltech-6V, and MELD. Its eight-dataset average ECE is $4.60\%$, reducing the $10.93\%$ obtained by UVF, the strongest baseline by average ECE, by 6.33 percentage points.

\subsubsection{Analysis}
\label{sec:ablation}

\paragraph*{Component analysis.}
Table~\ref{tab:ablation_components} compares classification results under different component settings. MRSL achieves a four-dataset mean ACC of $92.42\%$. We make the following observations:

\begin{table}[t]
    \centering
    \footnotesize
    \setlength{\tabcolsep}{0.5pt}
    \renewcommand{\arraystretch}{1.22}
    \caption{Component ablation of MRSL .}
    \label{tab:ablation_components}
    \begin{tabular*}{\linewidth}{@{\extracolsep{\fill}}lcccc@{}}
        \toprule
        Variant & Scene & Prokaryotic & Caltech-6V & Leaves100 \\
        \midrule
        Single-class & $82.32\!\pm\!1.16$ & $\underline{91.35\!\pm\!3.09}$ & $95.40\!\pm\!0.84$ & $99.56\!\pm\!0.36$ \\
        Random & $\underline{82.34\!\pm\!1.32}$ & $90.99\!\pm\!2.78$ & $95.65\!\pm\!0.87$ & $99.50\!\pm\!0.36$ \\
        Shared union & $81.94\!\pm\!1.36$ & $90.45\!\pm\!3.03$ & $\underline{95.69\!\pm\!0.67}$ & $\underline{99.62\!\pm\!0.26}$ \\
        Mean fusion & $79.96\!\pm\!1.06$ & $90.09\!\pm\!3.31$ & $\underline{95.69\!\pm\!0.28}$ & $98.44\!\pm\!0.73$ \\
        \midrule
        \textbf{MRSL} & $\mathbf{82.50\!\pm\!1.22}$ & $\mathbf{91.53\!\pm\!2.43}$ & $\mathbf{95.90\!\pm\!0.35}$ & $\mathbf{99.75\!\pm\!0.14}$ \\
        \bottomrule
    \end{tabular*}
\end{table}

\textbf{(1) Composite-class outputs benefit classification.} Retaining only single-class outputs (Single-class) reduces mean ACC on all four datasets, supporting the performance contribution of composite-class representations under the current experimental settings.

\textbf{(2) View-specific overlap-based selection outperforms the alternatives.} Random selection (Random) uniformly samples the same number of composite classes without replacement from class pairs with positive overlap scores. Shared union selection (Shared union) uses the deduplicated union of the per-view selections as a common output space, while masses are still learned independently per view. MRSL outperforms both schemes on all four datasets, supporting the effectiveness of selecting composite classes according to view-specific overlap structures.

\textbf{(3) Random-set fusion outperforms mass averaging.} We replace random-set fusion with mean fusion during both training and testing, averaging aligned subset masses before conversion to class probabilities. The four-dataset mean ACC decreases from $92.42\%$ to $91.04\%$, indicating that the adopted random-set fusion achieves better classification performance under the current settings.

\begin{figure}[t]
    \centering
    \includegraphics[width=\linewidth]{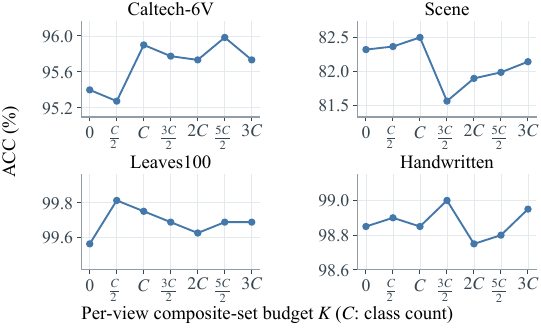}
    \caption{Effect of composite-class budget $K$ on ACC.}
    \label{fig:composite_budget}
\end{figure}

\paragraph*{Number of composite classes.}
Each view adds binary composite classes in descending order of training-set class-pair overlap scores. We vary the budget $K$ from $0$ to $3C$ in steps of $C/2$, rounding fractional budgets down. Figure~\ref{fig:composite_budget} shows that an appropriate budget improves mean ACC over single-class outputs on all four datasets, but the gains are non-monotonic in $K$. A larger candidate space therefore does not always yield higher accuracy.


\section{Conclusion}
\label{sec:conclusion}

In this paper, we proposed Multi-view Random-Set Learning (MRSL) to address view degradation in trusted multi-view learning. MRSL represents individual classes and ambiguous class groups through view-specific random sets for cross-view aggregation. For candidate-set selection, MRSL quantifies view-specific class overlap and constructs compact output spaces. For fusion, it combines view-specific mass assignments through set intersections, enabling complementary views to narrow candidate classes while retaining unresolved ambiguity. Experiments on synthetic and eight real-world datasets validate the overlap measure and demonstrate improvements in classification accuracy and confidence calibration.

\section*{LLM Usage Statement}

ChatGPT (OpenAI) was used for language polishing
of the manuscript, reference checking, and refinement
of schematic figures. Codex (OpenAI) was used to
assist with code development.

{
\small
\bibliographystyle{ieeenat_fullname}
\bibliography{main}
}

\clearpage
\setcounter{dbltopnumber}{1}
\appendix
\maketitlesupplementary

\setcounter{figure}{0}
\setcounter{table}{0}
\renewcommand{\thefigure}{S\arabic{figure}}
\renewcommand{\thetable}{S\arabic{table}}
\setcounter{equation}{0}
\renewcommand{\theequation}{S\arabic{equation}}
\renewcommand{\theHfigure}{supp.\arabic{figure}}
\renewcommand{\theHtable}{supp.\arabic{table}}
\renewcommand{\theHequation}{supp.\arabic{equation}}

This supplementary material provides further methodological and evaluation details for MRSL, together with additional experiments. Section~\ref{app:theoretical_background} describes overlap computation, Bayes reference values, and the definitions of uncertainty and expected calibration error (ECE). Section~\ref{app:experimental_settings} presents the datasets, compared methods, implementation settings, and synthetic data generation. Section~\ref{app:additional_results} uses controlled synthetic experiments, real-data overlap analysis, and calibration comparisons under a common confidence definition to further examine the validity of the overlap measure, MRSL's ability to handle view degradation, and its calibration performance.

\section{Method and evaluation details}
\label{app:theoretical_background}

This section details overlap computation, explains the Bayes reference curves, and defines evidential uncertainty and ECE.

\subsection{Overlap score computation}
\label{app:overlap_computation_details}

\textbf{Maximum-flow computation.} For a class pair $(a,b)$ in view $v$, let $n_a$ and $n_b$ denote the class sizes and $\mathcal{E}_{ab}^{(v)}$ the allowed cross-class connections. We construct a flow network following source $\rightarrow$ class-$a$ samples $\rightarrow$ class-$b$ samples $\rightarrow$ sink, where edge flows represent matching amounts. Source edges have capacity $1/n_a$, and sink edges have capacity $1/n_b$. A cross-class edge exists only when $(i,j)\in\mathcal{E}_{ab}^{(v)}$ and imposes no additional binding capacity constraint. Thus, each sample's total matching amount cannot exceed its capacity, and each class has total capacity one.

Let $\pi_{ij}$ denote the matching amount on cross-class edge $(i,j)$. The total network flow equals the sum of all cross-class matching amounts, so its maximum gives the class-pair overlap score:
\[
    O_{ab}^{(v)}
    =
    \max_{\pi\geq0}
    \sum_{(i,j)\in\mathcal{E}_{ab}^{(v)}}\pi_{ij},
\]
subject to the connection and capacity constraints above. We solve this problem using Dinic's algorithm~\cite{dinic1970algorithm}, obtaining maximum flow $O_{ab}^{(v)}$.

\textbf{Balanced participation allocation.} The same maximum matching total may admit different allocations across samples. To avoid dependence on the solver's allocation order, we preserve $O_{ab}^{(v)}$ and minimize
\[
    \frac{
        \displaystyle\sum_{i:y_i=a}
        \bigl(s_{ib}^{(v)}(\pi)\bigr)^2
    }{n_a}
    +
    \frac{
        \displaystyle\sum_{j:y_j=b}
        \bigl(s_{ja}^{(v)}(\pi)\bigr)^2
    }{n_b},
\]
under the original connection and capacity constraints. Here, $s_{ib}^{(v)}(\pi)$ is the fraction of sample $i$'s capacity used in matching with class $b$, and $s_{ja}^{(v)}(\pi)$ is defined analogously. This objective balances participation within each class as far as the connection structure permits.

After computing a maximum flow, its total value is fixed, but its allocation across samples may still be adjusted. Unused edge capacities represent possible flow increases, while reverse edges allow existing flow to be withdrawn and reassigned. These remaining forward capacities and reversible flows form the \emph{residual network}.

We use the residual network to identify parts in which matching amounts can be redistributed and balance them separately. Below, we consider a part where class-$b$ samples have fixed matching amounts equal to their full capacities, while participation among class-$a$ samples remains adjustable.

\textit{Checking uniform allocation.}
We divide the fixed matching total of the current part equally among its class-$a$ samples and use these amounts as matching limits. The checking network follows source $\rightarrow$ class-$b$ samples $\rightarrow$ class-$a$ samples $\rightarrow$ sink. Source-edge capacities equal the fixed matching amounts of class-$b$ samples, cross-class edges retain the original matching relations, and sink-edge capacities equal the current limits. If the maximum flow reaches the fixed total, the uniform allocation is feasible and this part is complete.

\textit{Adjusting matching limits.}
If the check fails, some samples must carry more than the initial uniform amount. In the residual network of the checking flow, we search from the source along edges with positive residual capacity. Reachable nodes and the remaining nodes define a minimum cut, identifying a constrained class-$b$ group and all its available class-$a$ neighbors. The group's fixed matching amount divided by its neighbors' total capacity gives the required capacity usage fraction. We increase the common participation limit accordingly and recompute the maximum flow, repeating this adjustment until the fixed total can be attained.

\textit{Fixing and rebalancing.}
Once the adjusted check succeeds, we fix participation in the last identified constrained group and remove its samples and corresponding matching amounts from the current part. We then redistribute the remaining matching amount uniformly among the remaining samples and repeat the feasibility check and any necessary adjustments. A feasible uniform allocation completes the remaining part; otherwise, further constrained groups are fixed until no samples remain to be processed. Class-$b$ participation is balanced symmetrically by exchanging the two sides. Parts with participation already fixed on both sides are retained directly.

The procedure preserves the maximum matching total. Strict convexity of the objective in participation ensures unique sample participation values, although individual edge flows need not be unique.

\textbf{Rank-weight derivation.} For each sample, we sort its participation relative to the other $C-1$ classes in descending order, denoted by $s_{i,(r)}^{(v)}$, where $C$ is the number of classes and $r$ is the rank. To emphasize stronger class overlap while retaining information about other competing classes, we choose decreasing base weights $1/[r(r+1)]$ and normalize them.

Using
\[
    \frac{1}{r(r+1)}
    =
    \frac{1}{r}-\frac{1}{r+1},
\]
we obtain
\[
    \sum_{r=1}^{C-1}\frac{1}{r(r+1)}
    =
    1-\frac{1}{C}
    =
    \frac{C-1}{C}.
\]
The normalized weights are therefore
\[
    w_r
    =
    \frac{C}{(C-1)r(r+1)},
    \qquad
    \sum_{r=1}^{C-1}w_r=1.
\]
The sample-level overlap score is
\[
    s_i^{(v)}
    =
    \sum_{r=1}^{C-1}
    w_r s_{i,(r)}^{(v)}.
\]

These weights are positive and decrease with rank, assigning greater weight to larger participation values. Since each participation value lies in $[0,1]$, the weighted sample score also lies in $[0,1]$. Finally, we average sample scores within each class and then equally across classes to obtain $G^{(v)}$, preventing larger classes from dominating the overall score.

\subsection{Bayes error and reference curves}
\label{app:bayes_reference}

This section describes the computation of Bayes reference curves for the two-view and single-view experiments. The two reference curves in the main paper are computed from the joint Gaussian distributions under the same-pair and complementary conditions and compared with the joint-space overlap score $G^{(3)}$. The single-view reference curve is computed from the two-dimensional class distributions and compared with the single-view overlap score $G$. We first describe the two-view case.

Given the class, the two views are conditionally independent, each with covariance $I_2$. Their concatenation therefore follows a four-dimensional Gaussian distribution with covariance $I_4$, the four-dimensional identity matrix. This means that, conditional on the class, all four coordinates have unit variance and are independent. Let $\mathbf{x}$ denote a concatenated feature location and $\boldsymbol{\mu}_c(q)$ the concatenated mean of class $c$. The class-conditional density is
\[
    f_c(\mathbf{x};q)
    =
    \frac{1}{(2\pi)^2}
    \exp\!\left(
        -\frac{
            \|\mathbf{x}-\boldsymbol{\mu}_c(q)\|^2
        }{2}
    \right).
\]
The same-pair and complementary conditions use their respective joint class means. The parameter $q$ controls overlap through these means, while the within-class spread remains fixed.

Since the three classes have equal priors and covariance, the Bayes-optimal rule selects the nearest class mean. For distinct means, the decision region of class $c$ is
\[
\begin{aligned}
    \mathcal{V}_c(q)=\bigl\{\mathbf{x}:\;
    &\|\mathbf{x}-\boldsymbol{\mu}_c(q)\|^2\\
    &\le
    \|\mathbf{x}-\boldsymbol{\mu}_j(q)\|^2,
    \ \forall j\ne c
    \bigr\}.
\end{aligned}
\]
Distance ties are resolved using a fixed rule. An observation generated from class $c$ is correctly classified when it falls in this region, giving the class-specific theoretical accuracy
\[
    A_c(q)
    =
    \int_{\mathcal{V}_c(q)}
    f_c(\mathbf{x};q)\,\mathrm{d}\mathbf{x}.
\]

The three classes are equally likely, so the Bayes error is
\[
    R^*(q)
    =
    1-\frac13\sum_{c=1}^{3}A_c(q).
\]
Since each location is assigned to a class with the largest class-conditional density, combining the decision regions also expresses the overall accuracy as an integral of the maximum class-conditional density. The normalized reference is
\[
    \widetilde{R}^*(q)
    =
    \frac{R^*(q)}{2/3}
    =
    \frac32\left[
        1-\frac13\int
        \max_c f_c(\mathbf{x};q)\,\mathrm{d}\mathbf{x}
    \right].
\]
Here, $R^*(q)$ is the Bayes error, and $\widetilde{R}^*(q)$ is the normalized reference used in the figure. In general, $C$ equally likely classes with identical distributions have optimal error $(C-1)/C$. Our setting uses $C=3$, giving the denominator $2/3$. Table~\ref{tab:bayes_reference_cases} illustrates three idealized cases.

\begin{table}[t]
    \centering
    \small
    \setlength{\tabcolsep}{2pt}
    \renewcommand{\arraystretch}{1.15}
    \caption{Idealized cases with three equally likely classes.}
    \label{tab:bayes_reference_cases}
    \begin{tabular*}{\linewidth}
        {@{\extracolsep{\fill}}lccc@{\hspace{10pt}}}
        \toprule
        Case & ACC & $R^*$ & $\widetilde{R}^*$ \\
        \midrule
        All separable
        & $1$ & $0$ & $0$ \\
        Two coincide; third separable
        & $2/3$ & $1/3$ & $1/2$ \\
        All coincide
        & $1/3$ & $2/3$ & $1$ \\
        \bottomrule
    \end{tabular*}
\end{table}

In the single-view case, each class follows a two-dimensional Gaussian distribution with covariance $I_2$. We apply the same decision rule to compute the class-specific theoretical accuracies, then obtain the Bayes error and its normalized reference as above.

\subsection{Evidential uncertainty}
\label{app:evidence_uncertainty}

In a standard Dirichlet evidential model, the model produces non-negative class evidence $e_{i,c}$ for sample $i$ and constructs Dirichlet parameters
\[
    \alpha_{i,c}=e_{i,c}+1,
    \qquad
    S_i=\sum_{c=1}^{C}\alpha_{i,c},
\]
where $C$ is the number of classes and $S_i$ is the total strength. The corresponding class probability and evidential uncertainty are
\[
    p_{i,c}=\frac{\alpha_{i,c}}{S_i},
    \qquad
    u_i=\frac{C}{S_i}.
\]
When all class evidence is zero, $u_i=1$. As the total evidence increases, $u_i$ decreases, reflecting uncertainty associated with insufficient evidence.

In the main-paper experiments, the eight trusted baselines compute $u_i$ from their fused outputs according to their respective definitions and use $1-u_i$ as predictive confidence.

\subsection{Calibration evaluation}
\label{app:ece_confidence_definitions}

Expected calibration error (ECE) measures calibration by comparing predictive confidence with empirical accuracy~\cite{guo2017calibration}. For $N$ test samples, let $\hat{y}_i=\arg\max_c p_{i,c}$ be the predicted class and $\gamma_i$ the confidence. We divide $[0,1]$ into 15 equal-width bins and denote the samples in bin $b$ by $B_b$. Its accuracy and average confidence are
\[
\begin{array}{@{}l@{}}
    \displaystyle
    \operatorname{acc}(B_b)
    = \frac{1}{|B_b|}
      \sum_{i\in B_b}\mathbb{I}(\hat y_i=y_i),
    \\[6pt]
    \displaystyle
    \operatorname{conf}(B_b)
    = \frac{1}{|B_b|}
      \sum_{i\in B_b}\gamma_i,
\end{array}
\]
where $\mathbb{I}(\cdot)$ is the indicator function. ECE is the weighted average of their absolute differences over non-empty bins:
\[
    \operatorname{ECE}
    =
    \sum_{b=1}^{15}
    \frac{|B_b|}{N}
    \left|
        \operatorname{acc}(B_b)
        -
        \operatorname{conf}(B_b)
    \right|.
\]
Lower ECE indicates closer agreement between confidence and empirical accuracy.

We use two confidence definitions:
\[
    \gamma_i^{\mathrm{evidence}}=1-u_i,
    \qquad
    \gamma_i^{\mathrm{probability}}=\max_c p_{i,c}.
\]
In the main-paper experiments, the eight trusted baselines use the former, while the remaining methods, including MRSL, use the latter. Appendix~\ref{app:additional_calibration} reports complete ECE results using the maximum predicted class probability as confidence for all methods. Both settings use the same bin boundaries and ECE formula, changing only the confidence definition while leaving predicted probabilities and classes unchanged.

\section{Experimental settings}
\label{app:experimental_settings}

This section introduces the datasets and compared methods, describes the data splits, preprocessing, and model training settings, and details the synthetic data generation procedures.

\subsection{Dataset details}
\label{app:dataset_details}

\begin{table}[t]
    \centering
    \small
    \setlength{\tabcolsep}{3pt}
    \renewcommand{\arraystretch}{1.12}
    \begin{tabular}{@{}lrrl@{}}
        \toprule
        Dataset & Size & $C$ & Dimensionality \\
        \midrule
        Handwritten & 2,000 & 10 & 240/76/216/47/64/6 \\
        Scene & 4,485 & 15 & 20/59/40 \\
        Caltech-6V & 2,386 & 20 & 48/40/254/1984/512/928 \\
        MSRCV & 210 & 7 & 1302/48/512/100/256/210 \\
        Leaves100 & 1,600 & 100 & 64/64/64 \\
        Prokaryotic & 551 & 4 & 438/3/393 \\
        CUB & 600 & 10 & 1024/300 \\
        MELD & 12,596 & 3 & 300/1422 \\
        \bottomrule
    \end{tabular}
    \caption{Dataset summary.}
    \label{tab:dataset_statistics}
\end{table}

\textbf{Handwritten}\footnote{\url{https://archive.ics.uci.edu/dataset/72/multiple+features}} contains 2,000 handwritten digit samples from ten classes, represented by six pre-extracted feature views covering pixel statistics, transform coefficients, and shape descriptors.

\textbf{Scene}\footnote{\url{https://slazebni.cs.illinois.edu/research/scene_categories.zip}} is a scene recognition dataset containing 4,485 images from 15 indoor and outdoor categories, represented by three pre-extracted feature views: GIST, PHOG, and LBP.

\textbf{Caltech-6V}\footnote{\url{https://data.caltech.edu/records/mzrjq-6wc02}} is a 20-class subset of Caltech-101 containing 2,386 images, represented by six feature views: Gabor, wavelet moments, CENTRIST, HOG, GIST, and LBP.

\textbf{MSRCV}\footnote{\url{https://download.microsoft.com/download/A/1/1/A116CD80-5B79-407E-B5CE-3D5C6ED8B0D5/msrc_objcategimagedatabase_v1.zip}} is a seven-class image classification subset of MSRC-v1 containing 210 images, each represented by six pre-extracted visual feature views.

\begin{table*}[t]
    \centering
    \caption{Main design characteristics of the compared methods. $\checkmark$ and -- indicate whether each design is explicitly included. Adaptive weighting includes view-contribution weighting and trust discounting. Composite sets refer to sets containing multiple classes.}
    \label{tab:method_characteristics}
    \small
    \setlength{\tabcolsep}{3pt}
    \renewcommand{\arraystretch}{1.25}
    \begin{tabular*}{\textwidth}{@{\extracolsep{\fill}}l*{11}{c}>{\columncolor[gray]{0.94}}c@{}}
        \toprule
        & \multicolumn{8}{c}{\textbf{Evidence-based trusted methods}}
        & \multicolumn{3}{c}{\textbf{Adaptive fusion methods}}
        & \multicolumn{1}{>{\columncolor[gray]{0.94}}c}{\textbf{Ours}} \\
        \cmidrule(lr){2-9}
        \cmidrule(lr){10-12}
        \cmidrule(lr){13-13}
        \textbf{Characteristic}
        & TMC & ECML & CCML & TMCEK
        & RCMCL & TUNED & HOVQ & ETF
        & UVF & AssoDMVC & PDF & \textbf{MRSL} \\
        \midrule
        Evidential modeling
        & $\checkmark$ & $\checkmark$
        & $\checkmark$ & $\checkmark$
        & $\checkmark$ & $\checkmark$
        & $\checkmark$ & $\checkmark$
        & -- & -- & -- & $\checkmark$ \\
        Adaptive weighting
        & -- & -- & -- & --
        & $\checkmark$ & $\checkmark$
        & -- & $\checkmark$
        & $\checkmark$ & $\checkmark$
        & $\checkmark$ & -- \\
        Vagueness modeling
        & -- & -- & $\checkmark$ & --
        & -- & -- & $\checkmark$ & --
        & -- & -- & -- & $\checkmark$ \\
        Composite-set modeling
        & -- & -- & -- & --
        & -- & -- & $\checkmark$ & --
        & -- & -- & -- & $\checkmark$ \\
        \bottomrule
    \end{tabular*}
\end{table*}

\textbf{Leaves100}\footnote{\url{https://archive.ics.uci.edu/dataset/241/one+hundred+plant+species+leaves+data+set}} is a plant leaf classification dataset containing 1,600 samples from 100 species, represented by three feature views: shape, margin, and texture.

\textbf{Prokaryotic}\footnote{\url{https://github.com/mbrbic/Multi-view-LRSSC}} is a prokaryotic species classification dataset containing 551 species from four phyla, represented by three views: text, proteome composition, and gene repertoire.

\textbf{CUB}\footnote{\url{https://www.vision.caltech.edu/datasets/cub_200_2011/}} is a ten-class subset of Caltech-UCSD Birds containing 600 samples, represented by image and text feature views for bird classification.

\textbf{MELD}\footnote{\url{https://affective-meld.github.io/}} is a multimodal conversational sentiment dataset. We use the neutral, positive, and negative three-class task with text and audio views, retaining 9,986 training and 2,610 test samples from the official split.

\subsection{Compared methods}
\label{app:compared_methods}

We compare MRSL with two categories of representative multi-view classification methods. The first comprises \emph{evidence-based trusted multi-view learning methods}. \textbf{TMC}~\cite{han2021trusted} models class evidence and uncertainty using Dirichlet distributions and combines opinions through the Dempster--Shafer rule. \textbf{ECML}~\cite{xu2024ecml} introduces evidence aggregation and conflict constraints. \textbf{CCML}~\cite{liu2024dynamic} decouples consistent and complementary evidence, relaxing supervision on the latter to accommodate semantic vagueness. \textbf{TMCEK}~\cite{liang2025tmcek} incorporates expert knowledge constraints and evidence-distribution information. \textbf{RCMCL}~\cite{hu2026rcmcl} addresses view conflicts through evidential contrastive and collaborative learning. \textbf{TUNED}~\cite{huang2025tuned} models local and global feature-neighborhood structures and selectively aggregates view evidence. \textbf{HOVQ}~\cite{hu2026hyperopinion} quantifies vagueness using hyper-opinions and guides training through gradient modulation. \textbf{ETF}~\cite{lu2025trust} estimates prediction trustworthiness and discounts unreliable opinions before fusion.

The second category comprises \emph{adaptive feature utilization and dynamic fusion methods}. \textbf{UVF}~\cite{lv2026uvf} selects features according to view strength and adjusts fusion weights using uncertainty. \textbf{AssoDMVC}~\cite{liang2025assodmvc} models inter-view associations to guide feature modulation and dynamic fusion. \textbf{PDF}~\cite{cao2024pdf} predicts view contributions and relatively calibrates the resulting fusion weights.

Table~\ref{tab:method_characteristics} summarizes their main characteristics. Although HOVQ and MRSL both use composite sets, HOVQ projects composite-set support onto singleton classes before fusion. MRSL instead selects composite sets from each view's training-data class overlap and preserves their support during cross-view fusion, allowing complementary views to refine candidate classes through set intersections.

\subsection{Implementation details}
\label{app:implementation_details}

For the first seven datasets, a stratified 80/20 train/test split is generated separately for each seed, and all methods share the same split for that seed. MELD uses its fixed official train/test split across all seeds. Each view is independently Min--Max normalized, with parameters fitted only on training data and then applied to test data.

The reference environment runs Ubuntu 22.04.4 LTS with Linux kernel 5.15.0-105-generic, an Intel Xeon Gold 6142 CPU (2.60 GHz), 24 GiB of system memory, and an NVIDIA GeForce RTX 3080 GPU with 10 GB memory. The CUDA runtime and NVIDIA driver versions are 12.4 and 610.57.04, respectively. The implementation uses Python 3.12.3, PyTorch 2.5.1, NumPy 2.1.3, SciPy 1.18.1, and scikit-learn 1.9.1.

Each view in MRSL uses an independent fully connected encoder with hidden width 256, followed by ReLU, Dropout, and BatchNorm. A Softmax output layer produces mass assignments over singleton classes and composite sets. Composite sets are constructed and selected independently using only each view's training features and labels, and remain fixed throughout training.

MRSL is trained for 300 epochs using Adam with a batch size of 200 and an initial learning rate of $10^{-3}$, with linear warm-up for the first 10 epochs followed by cosine decay. The auxiliary loss weight is fixed at $\lambda=0.3$, and the per-view composite-set budget is $K=C$, where $C$ is the number of classes. We use five seeds, $\{3407,7,601,101,503\}$, without early stopping, and evaluate the final model at epoch 300 to report ACC and ECE~\cite{guo2017calibration}.

Results are reported as mean $\pm$ standard deviation over five runs. ECE uses 15 equal-width bins. For the ECE results reported in the main paper, the eight trusted baselines use confidence $1-u$, while the remaining methods, including MRSL, use the maximum predicted class probability. The definitions of uncertainty and ECE are provided in Sections~\ref{app:evidence_uncertainty} and~\ref{app:ece_confidence_definitions}, respectively. Section~\ref{app:additional_calibration} further reports complete ECE results using the maximum predicted class probability as confidence for all methods.

\subsection{Synthetic data generation}
\label{app:synthetic_data_generation}

This section describes two types of synthetic data used to validate the overlap measure and analyze set representation and classification performance, respectively. We further examine performance under view degradation by progressively weakening discriminative information.

\subsubsection{Overlap-validation data}
\label{app:overlap_validation_data}

The main paper uses two-view Gaussian data to examine how $G$ responds to cross-view complementarity. The data contain three equally likely classes, with 400 samples per class and covariance $I_2$, the two-dimensional identity matrix. For two classes whose means are distance $d$ apart, the density overlap coefficient satisfies
\[
    q=2\Phi(-d/2),
    \qquad
    d(q)=2\Phi^{-1}(1-q/2),
\]
where $\Phi$ is the standard normal cumulative distribution function. We use $q\in\{0.01,0.25,0.50,0.75,1\}$, with $q=0.01$ representing nearly separated classes to avoid the infinite separation required for zero overlap.

\begin{figure}[t]
    \centering
    \includegraphics[width=\linewidth]{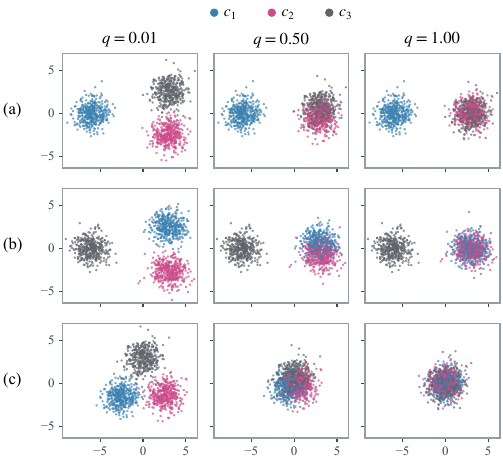}
    \caption{Overlap-validation data: (a) view~1 shared by both two-view conditions; (b) view~2 under the complementary condition; (c) the single-view setting.}
    \label{fig:app_synthetic_overlap_data}
\end{figure}

In view~1, shown in Fig.~\ref{fig:app_synthetic_overlap_data}(a), the class means are respectively
\[
    (-6,0),\qquad
    (3,-d/2),\qquad
    (3,d/2),
\]
where $d=d(q)$. As $q$ increases, $c_2$ and $c_3$ increasingly overlap. Under the \emph{same-pair condition}, view~2 uses the same means with independent sampling. Under the \emph{complementary condition} shown in Fig.~\ref{fig:app_synthetic_overlap_data}(b), we exchange the $c_1$ and $c_3$ sample blocks in view~2, changing its overlapping pair to $c_1$ and $c_2$. Both conditions preserve the single-view overlap structure and change only the cross-view class correspondence.

We additionally use the single-view setting in Fig.~\ref{fig:app_synthetic_overlap_data}(c) to examine how $G$ responds to class overlap itself. The three class means lie at the vertices of an equilateral triangle with side length $d(q)$, while the remaining sampling settings are unchanged, giving every class pair overlap coefficient $q$. Results are reported in Appendix~\ref{app:controlled_synthetic_experiments}, with normalized Bayes error calculated as described in Appendix~\ref{app:bayes_reference}.

All settings use five data-generation seeds. Within each seed, only the means change across overlap levels, with Gaussian noise realizations reused. Before computing $G$, each view is standardized feature-wise and divided by $\sqrt{2}$; joint features concatenate the views. All 1,200 samples are used directly for overlap computation without training a classifier.

\subsubsection{Set-representation and classification data}
\label{app:synthetic_classification_data}

We construct a two-view dataset with three classes and 400 samples per class. In the main-paper setting, view~1 cannot distinguish $c_2$ from $c_3$, while view~2 cannot distinguish $c_1$ from $c_2$, allowing us to examine candidate-group representation and the use of complementary information.

Each nine-dimensional latent vector $\boldsymbol{\xi}$ contains three three-dimensional blocks: $c_1$ activates only the first block, $c_2$ only the second, and $c_3$ both the second and third. Active coordinates are independently sampled from
\[
    \operatorname{Gamma}(1,0.9)
    +\mathcal{N}(0,0.1^2)+0.5,
\]
where the Gamma parameters specify shape and scale; all remaining coordinates are zero. Two 12-dimensional views are generated as
\[
    \boldsymbol{x}^{(v)}
    =
    W^{(v)}\boldsymbol{\xi}
    +\boldsymbol{\eta}^{(v)}.
\]

First, entries of $W^{(v)}\in\mathbb{R}^{12\times9}$ are sampled from $\mathcal{U}(0.4,1.0)$ and independently set to zero with probability $0.3$. We then zero the last three columns of $W^{(1)}$ and the first six columns of $W^{(2)}$, removing information that distinguishes $c_2/c_3$ and $c_1/c_2$, respectively, to obtain the complementary-view setting used in the main paper. Observation noise follows
\[
    \boldsymbol{\eta}^{(1)}
    \sim \mathcal{N}(\boldsymbol{0},\,0.5^2I_{12}),
    \qquad
    \boldsymbol{\eta}^{(2)}
    \sim \mathcal{N}(\boldsymbol{0},\,I_{12}).
\]

To examine performance under different degrees of view degradation, we retain the random element-wise zeroing and replace the subsequent zeroing of designated columns with multiplication by $1-t$, where $t\in\{0,0.25,0.50,0.75,1\}$. Increasing $t$ progressively weakens the information distinguishing $c_2$ from $c_3$ in view~1 and $c_1$ from $c_2$ in view~2, making independent class discrimination more difficult for each view. At $t=0$, the corresponding information is retained; $t=1$ recovers the fully masked main-paper setting. Figure~\ref{fig:app_masked_synthetic_data} shows the resulting inputs, with results reported in Appendix~\ref{app:controlled_synthetic_experiments}.

All masking levels share the same latent samples, randomly initialized mappings, observation noise, and stratified 80/20 train/test split. Min--Max normalization is fitted separately on each view's training features and applied to its test features. The five runs use identical data and partitions, varying only the training random seed.

\begin{figure}[t]
    \centering
    \includegraphics[width=\linewidth]{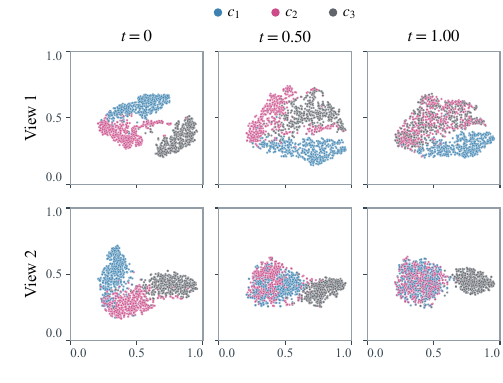}
    \caption{t-SNE visualization of synthetic inputs at different masking strengths. Rows correspond to views~1 and~2; $t=1$ corresponds to the main-paper setting.}
    \label{fig:app_masked_synthetic_data}
\end{figure}

\section{Additional experimental results}
\label{app:additional_results}

This section further examines the validity of the overlap measure and MRSL's ability to handle view degradation. First, synthetic experiments assess how the overlap score responds to class overlap and compare classification and calibration as single-view discriminative information is progressively weakened. We then compare classification performance across different levels of single-view overlap on real-world datasets. Finally, we compare calibration using the maximum predicted class probability as confidence for all methods.

\subsection{Controlled synthetic experiments}
\label{app:controlled_synthetic_experiments}

We first provide additional single-view overlap validation, then examine classification and calibration under view degradation.

\begin{figure}[t]
    \centering
    \includegraphics[width=\linewidth]{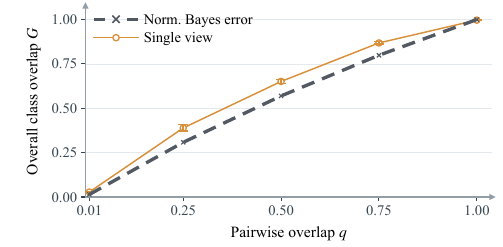}
    \caption{Single-view overlap validation.}
    \label{fig:app_single_view_overlap}
\end{figure}

\paragraph*{$G$ increases with class overlap.}
Figure~\ref{fig:app_single_view_overlap} compares $G$ with normalized Bayes error at five overlap levels $q\in\{0.01,0.25,0.50,0.75,1\}$, using all 1,200 Gaussian samples defined in Appendix~\ref{app:overlap_validation_data} and reporting mean $\pm$ standard deviation over five data-generation seeds. As $q$ increases, $G$ rises from approximately $0.03$ to nearly $1$, following the theoretical reference described in Appendix~\ref{app:bayes_reference}. Greater overlap allows more participation in cross-class matching, enabling $G$ to reflect the transition from nearly separated to strongly overlapping class distributions.

\paragraph*{View degradation.}
To examine performance as single-view discriminative information weakens, we use the five masking strengths defined in Appendix~\ref{app:synthetic_classification_data}, progressively weakening the information distinguishing $c_2/c_3$ in view~1 and $c_1/c_2$ in view~2. CCML, HOVQ, and MRSL use the same 64-dimensional encoders with their respective losses and fusion mechanisms, and results are reported as mean $\pm$ standard deviation over five training seeds. Whereas CCML and HOVQ use $1-u$ as confidence in the main paper, this experiment uses the maximum predicted class probability for all methods. Overlap scores are computed only from training data, with $\bar{G}=(G^{(1)}+G^{(2)})/2$ denoting mean single-view overlap and $G^{(3)}$ denoting joint-space overlap.

\begin{table*}[t]
    \centering
    \small
    \setlength{\tabcolsep}{3pt}
    \renewcommand{\arraystretch}{1.20}
    \caption{Classification and calibration at five masking strengths (ACC/ECE in \%, mean $\pm$ std.). Best means are bold.}
    \label{tab:app_masking_results}
    \begin{tabular*}{\textwidth}{@{\extracolsep{\fill}}rrr rr rr rr@{}}
        \toprule
        \multicolumn{3}{c}{Data setting}
        & \multicolumn{2}{c}{CCML}
        & \multicolumn{2}{c}{HOVQ}
        & \multicolumn{2}{c}{MRSL} \\
        \cmidrule(lr){1-3}
        \cmidrule(lr){4-5}
        \cmidrule(lr){6-7}
        \cmidrule(lr){8-9}
        \multicolumn{1}{c}{$t$}
        & \multicolumn{1}{c}{$\bar{G}$}
        & \multicolumn{1}{c}{$G^{(3)}$}
        & \multicolumn{1}{c}{ACC$\uparrow$}
        & \multicolumn{1}{c}{ECE$\downarrow$}
        & \multicolumn{1}{c}{ACC$\uparrow$}
        & \multicolumn{1}{c}{ECE$\downarrow$}
        & \multicolumn{1}{c}{ACC$\uparrow$}
        & \multicolumn{1}{c}{ECE$\downarrow$} \\
        \midrule
        $0.00$ & $0.2425$ & $0.1269$
        & $99.25\pm0.54$ & $8.85\pm0.31$
        & $99.25\pm0.19$ & $33.67\pm0.29$
        & $\mathbf{99.50\pm0.19}$ & $\mathbf{0.79\pm0.15}$ \\
        $0.25$ & $0.3116$ & $0.1791$
        & $\mathbf{98.83\pm0.35}$ & $9.95\pm0.36$
        & $97.67\pm0.63$ & $33.43\pm0.69$
        & $\mathbf{98.83\pm0.54}$ & $\mathbf{1.25\pm0.32}$ \\
        $0.50$ & $0.4314$ & $0.2397$
        & $96.83\pm0.56$ & $11.54\pm0.67$
        & $95.92\pm0.35$ & $34.05\pm0.39$
        & $\mathbf{97.58\pm0.75}$ & $\mathbf{1.42\pm0.22}$ \\
        $0.75$ & $0.5653$ & $0.2954$
        & $92.58\pm1.36$ & $13.82\pm1.17$
        & $90.92\pm0.35$ & $33.14\pm0.58$
        & $\mathbf{96.83\pm0.56}$ & $\mathbf{2.48\pm0.98}$ \\
        $1.00$ & $0.6053$ & $0.3279$
        & $91.00\pm1.63$ & $18.47\pm1.55$
        & $90.08\pm1.12$ & $37.92\pm0.95$
        & $\mathbf{95.17\pm0.48}$ & $\mathbf{3.08\pm0.47}$ \\
        \bottomrule
    \end{tabular*}
\end{table*}

\textbf{Smaller accuracy declines.} Table~\ref{tab:app_masking_results} compares classification performance at five masking strengths alongside mean single-view and joint-space overlap. As masking increases, both overlap scores rise, but MRSL's ACC declines by only 4.33 percentage points from $t=0$ to $t=1$, compared with 8.25 for CCML and 9.17 for HOVQ; at $t=0.75$ and $t=1$, MRSL exceeds the strongest baseline by 4.25 and 4.17 percentage points, respectively. This pattern is consistent with their treatment of within-group ambiguity: CCML's class-wise complementary evidence may introduce support for incorrect classes, while HOVQ converts set support to singleton support before fusion and may retain unreliable within-group preferences. MRSL preserves candidate-group support until fusion, allowing information-insufficient views to provide useful candidate ranges that complementary views further refine.

\textbf{Lower ECE under a common definition.} Table~\ref{tab:app_masking_results} compares calibration across masking strengths using maximum predicted class probability as confidence for all three methods. MRSL achieves the lowest ECE at every level; at $t=1$, its ECE is $3.08\%$, compared with $18.47\%$ for CCML and $37.92\%$ for HOVQ. This may reflect differences in evidence learning and fusion: CCML's consistent evidence is limited by the lower-support view and its complementary evidence can spread support across competing classes, while HOVQ's entropy regularization may discourage evidence concentration and produce conservative confidence. MRSL retains candidate-group support and uses a classification objective on the fused output to exploit joint discriminative information, allowing its final probabilities to adapt to cross-view information, consistent with the lower ECE observed here.

\subsection{View degradation analysis on real-world data}
\label{app:real_overlap_analysis}

We further analyze MRSL's ability to handle view degradation on real-world data. An individual view may struggle to distinguish certain classes independently, so we use single-view class overlap as a geometric characterization of insufficient discriminative information and compare classification performance across overlap levels.

For each seed, we normalize test features using parameters obtained from the training set and compute each sample's view-averaged overlap score on the complete test set, using its features and ground-truth labels as described in Section~\ref{app:overlap_computation_details}:
\[
    \bar{s}_i
    =
    \frac{1}{V}\sum_{v=1}^{V}s_i^{(v)},
\]
where $s_i^{(v)}$ is obtained by weighting sample $i$'s participation in overlaps with competing classes in view $v$, and $V$ is the number of views. Larger $\bar{s}_i$ indicates stronger average single-view class overlap. These scores are used only for post-hoc analysis and do not affect training or composite-set selection.

We divide $[0,1]$ into five fixed intervals of width $0.2$ and assign test samples according to $\bar{s}_i$. Within each interval, we pool the final prediction records from five seeds and compute ACC as the number of correct predictions divided by the total number of records.

\begin{figure}[t]
    \centering
    \includegraphics[width=\linewidth]{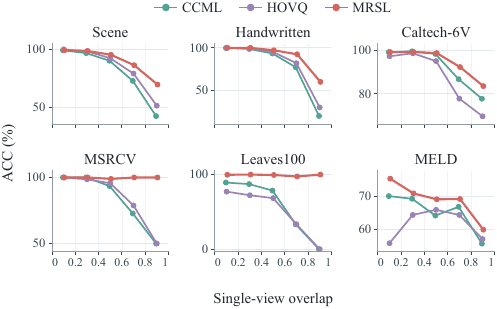}
    \caption{Classification accuracy across five fixed overlap intervals on six real-world datasets.}
    \label{fig:app_real_overlap_acc}
\end{figure}

\begin{figure}[t]
    \centering
    \includegraphics[width=\linewidth]{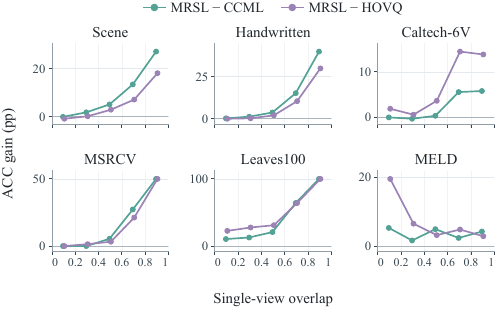}
    \caption{ACC gains of MRSL over CCML and HOVQ in percentage points.}
    \label{fig:app_real_overlap_gain}
\end{figure}

\begin{table*}[t]
    \centering
    \small
    \setlength{\tabcolsep}{2pt}
    \renewcommand{\arraystretch}{1.12}
    \caption{Calibration results using maximum predicted class probability as confidence (ECE in \%, mean $\pm$ standard deviation). Best and second-best means are bold and underlined.}
    \label{tab:app_ece_max_probability}
    \resizebox{\textwidth}{!}{%
    \begin{tabular}{@{}l*{8}{c}@{}}
        \toprule
        Model & Scene & Handwritten & Caltech-6V & MSRCV & Leaves100 & Prokaryotic & CUB & MELD \\
        \midrule
        TMC & $52.42\pm1.18$ & $20.22\pm1.19$ & $15.77\pm0.79$ & $31.42\pm2.25$ & $85.42\pm1.83$ & $38.35\pm1.51$ & $40.87\pm1.84$ & $10.57\pm0.58$ \\
        ECML & $55.60\pm1.41$ & $57.50\pm1.29$ & $50.89\pm0.88$ & $62.06\pm2.62$ & $90.24\pm1.32$ & $32.41\pm3.40$ & $45.65\pm2.54$ & $17.31\pm0.46$ \\
        CCML & $52.85\pm1.32$ & $49.42\pm1.12$ & $48.58\pm0.80$ & $49.65\pm2.12$ & $84.27\pm2.03$ & $29.06\pm2.20$ & $40.58\pm1.48$ & $9.51\pm0.46$ \\
        TMCEK & $59.07\pm1.06$ & $70.69\pm1.03$ & $59.88\pm0.87$ & $48.39\pm4.39$ & $90.70\pm1.53$ & $31.51\pm0.63$ & $52.10\pm2.28$ & $17.66\pm0.74$ \\
        RCMCL & $53.29\pm1.37$ & $35.78\pm1.03$ & $36.68\pm0.62$ & $42.27\pm1.39$ & $92.49\pm1.39$ & $26.07\pm1.22$ & $43.28\pm2.11$ & $\mathbf{6.38\pm2.24}$ \\
        TUNED & $58.52\pm1.56$ & $51.92\pm1.26$ & $58.55\pm2.95$ & $47.32\pm2.09$ & $86.42\pm2.00$ & $42.09\pm6.21$ & $54.90\pm3.02$ & $20.88\pm0.81$ \\
        HOVQ & $60.81\pm1.58$ & $58.99\pm1.09$ & $69.11\pm0.87$ & $57.44\pm2.57$ & $71.88\pm2.49$ & $29.92\pm1.16$ & $69.43\pm2.58$ & $17.07\pm0.45$ \\
        ETF & $53.76\pm0.69$ & $31.92\pm1.07$ & $21.25\pm0.79$ & $43.58\pm3.08$ & $94.97\pm0.81$ & $18.77\pm0.37$ & $46.94\pm1.72$ & $8.18\pm1.61$ \\
        UVF & $19.31\pm2.47$ & $\underline{1.13\pm0.49}$ & $\mathbf{2.16\pm0.94}$ & $\underline{10.50\pm4.69}$ & $\underline{1.27\pm0.25}$ & $\underline{13.33\pm3.04}$ & $\underline{5.19\pm1.52}$ & $34.51\pm4.07$ \\
        AssoDMVC & $\mathbf{7.72\pm0.69}$ & $8.72\pm2.65$ & $6.24\pm1.26$ & $22.62\pm2.62$ & $3.40\pm0.23$ & $13.52\pm0.46$ & $12.45\pm1.15$ & $35.63\pm0.79$ \\
        PDF & $15.93\pm1.74$ & $3.97\pm0.60$ & $3.56\pm0.75$ & $19.54\pm1.20$ & $39.27\pm0.69$ & $28.07\pm2.93$ & $7.13\pm1.52$ & $35.37\pm6.08$ \\
        \midrule
        MRSL & $\underline{12.09\pm1.16}$ & $\mathbf{1.04\pm0.46}$ & $\underline{3.54\pm0.38}$ & $\mathbf{0.36\pm0.70}$ & $\mathbf{0.27\pm0.15}$ & $\mathbf{7.43\pm1.89}$ & $\mathbf{4.37\pm1.31}$ & $\underline{7.68\pm2.40}$ \\
        \bottomrule
    \end{tabular}%
    }
\end{table*}

\textbf{Advantages on high-overlap samples.} Figure~\ref{fig:app_real_overlap_acc} shows classification accuracy across five fixed overlap intervals on six real-world datasets, while Fig.~\ref{fig:app_real_overlap_gain} shows MRSL's ACC gains over CCML and HOVQ. On Scene and Handwritten, accuracy generally decreases as overlap increases, but MRSL exhibits a smaller decline and larger relative gains; Caltech-6V also shows larger gains in higher-overlap intervals. For example, in Scene's highest-overlap interval, MRSL achieves $69.73\%$ ACC, exceeding CCML and HOVQ by $27.12$ and $18.15$ percentage points, respectively. MRSL also maintains high accuracy in higher-overlap intervals on MSRCV and Leaves100 and outperforms both baselines in every interval on MELD. These results are consistent with MRSL's approach to insufficient single-view information: when an individual view struggles to distinguish classes within a group, MRSL preserves the candidate range supported by that view and uses complementary information through cross-view fusion to refine discrimination, helping mitigate the effect of insufficient single-view discriminative information on final classification.

\subsection{Calibration with common confidence}
\label{app:additional_calibration}

To compare calibration under a common confidence definition, we compute ECE using the maximum predicted class probability, $\gamma_i=\max_c p_{i,c}$, for all methods. We retain the model checkpoints and test data used in the main paper, use 15 equal-width bins, and report mean $\pm$ standard deviation over five runs. The results for UVF, AssoDMVC, PDF, and MRSL remain unchanged, as these methods already use this confidence definition.

Table~\ref{tab:app_ece_max_probability} shows that MRSL achieves the lowest ECE on five datasets and ranks second on the remaining three. Its average ECE across the eight datasets is $4.60\%$, compared with $10.93\%$ for UVF, the strongest baseline by average ECE. These results show that MRSL retains its calibration advantage under a common confidence definition, supporting closer agreement between its maximum predicted class probabilities and empirical accuracy.

\end{document}